%% file: main.tex
\documentclass[sigconf]{acmart}

\input{misc/customize}

\def\BibTeX{{\rm B\kern-.05em{\sc i\kern-.025em b}\kern-.08emT\kern-.1667em\lower.7ex\hbox{E}\kern-.125emX}}
    
\copyrightyear{2020}
\acmYear{2020}
\setcopyright{acmlicensed}\acmConference[ICCAD '20]{IEEE/ACM International
Conference on Computer-Aided Design}{November 2--5, 2020}{Virtual Event,
USA}
\acmBooktitle{IEEE/ACM International Conference on Computer-Aided Design
(ICCAD '20), November 2--5, 2020, Virtual Event, USA}
\acmPrice{15.00}
\acmDOI{10.1145/3400302.3415685}
\acmISBN{978-1-4503-8026-3/20/11}

\begin{document}

%
% The "title" command has an optional parameter, allowing the author to define a "short title" to be used in page headers.
\title{A general approach for identifying hierarchical symmetry constraints for analog circuit layout}

%
% The "author" command and its associated commands are used to define the authors and their affiliations.
% Of note is the shared affiliation of the first two authors, and the "authornote" and "authornotemark" commands
% used to denote shared contribution to the research.
%\author{           $\; \; \; \; \; \; \; \;$ 
%Kishor Kunal       $\; \; \; \; \; \; \; \;$
%Jitesh Poojary     $\; \; \; \; \; \; \; \; \; \; \; \; \; \;$
%Tonmoy Dhar        $\; \; \; \; \; \; \; \;$
%Meghna Madhusudan  $\; \; \; \; \; \; \; \;$
%Ramesh Harjani     $\; \; \; \; \; \; \; \; \; $
%Sachin S. Sapatnekar}
%\normalsize{University of Minnesota, MN }}

%\orcid{1234-5678-9012}
%\affiliation{%
%\institution{University of Minnesota, Minneapolis, MN, USA}
  %\institution{Department of Electrical and Computer Engineering\\ University of Minnesota}
  %\streetaddress{200 Union St. SE}
  %\city{Minneapolis}
  %\state{MN (USA)}
  %\postcode{55455}
%}
%\email{sachin@umn.edu}

\author{Kishor Kunal}
\affiliation{\institution{University of Minnesota}}
\author{Jitesh Poojary}
\affiliation{\institution{University of Minnesota}}
\author{Tonmoy Dhar}
\affiliation{\institution{University of Minnesota}}
\author{Meghna Madhusudan}
\affiliation{\institution{University of Minnesota}}
\author{Ramesh Harjani}
\affiliation{\institution{University of Minnesota}}
\author{Sachin S. Sapatnekar}
\affiliation{\institution{University of Minnesota}}

%
% By default, the full list of authors will be used in the page headers. Often, this list is too long, and will overlap
% other information printed in the page headers. This command allows the author to define a more concise list
% of authors' names for this purpose.
\renewcommand{\shortauthors}
{K. Kunal, J. Poojary, T. Dhar, M. Madhusudan, R. Harjani, and S. S. Sapatnekar}

%
% The abstract is a short summary of the work to be presented in the article.

\begin{abstract}
Analog layout synthesis requires some elements in the circuit netlist to be
matched and placed symmetrically. However, the set of symmetries is very
circuit-specific and a versatile algorithm, applicable to a broad variety of
circuits, has been elusive.  This paper presents a general methodology for the
automated generation of symmetry constraints, and applies these constraints to
guide automated layout synthesis.  While prior approaches were restricted to
identifying simple symmetries, the proposed method operates hierarchically and
uses graph-based algorithms to extract multiple axes of symmetry within a
circuit.  An important ingredient of the algorithm is its ability to identify
arrays of repeated structures.  In some circuits, the repeated structures are
not perfect replicas and can only be found through approximate graph
matching.  A fast graph neural network based methodology is developed for this
purpose, based on evaluating the graph edit distance. The utility of this
algorithm is demonstrated on a variety of circuits, including operational
amplifiers, data converters, equalizers, and low-noise amplifiers.
\end{abstract}	

%
% The code below is generated by the tool at http://dl.acm.org/ccs.cfm.
% Please copy and paste the code instead of the example below.
%



%
% Keywords. The author(s) should pick words that accurately describe the work being
% presented. Separate the keywords with commas.

%
% This command processes the author and affiliation and title information and builds
% the first part of the formatted document.
\maketitle

\input{sec/1-intro}

\input{sec/2-method}

\input{sec/3-SimGNN}

\input{sec/4-results}
\input{sec/5-conclusion}

\begin{acks}
This work was supported in part by DARPA IDEA program under SPAWAR contract
N660011824048. 
\end{acks}

%
% The next two lines define the bibliography style to be used, and the bibliography file.
%\bibliographystyle{ACM-Reference-Format}
%\bibliography{bib/main}
\bibliographystyle{misc/ieeetr2}
\bibliography{bib/main-new}

\end{document}

%% file: misc/customize.tex
\usepackage{amsmath}
\usepackage{algpseudocode}
\usepackage{algorithm}
\usepackage{algcompatible}
\usepackage{subfigure}
\newcommand{\cal}[1]{\mathcal{#1}}
\newcommand{\ignore}[1]{}

\newcommand{\blueHL}[1]{{\textcolor{blue}{#1}}}

\usepackage{tikz}
\usetikzlibrary{shapes,arrows}

%% file: sec/1-intro.tex
\section{Introduction}
\label{sec:intro}

\noindent
Specialized layout techniques involving forms of symmetry, such as symmetry
about an axis, common-centroid layout, and matching, have long been used by
analog layout engineers to achieve high performance and high yield in analog
designs.  Matching techniques are important for both active and passive
elements~\cite{Pelgrom89}, which are subject to perturbations due to random and
systematic variations as well as changing operating conditions.  Analog
circuits convert the less controllable problem of reducing absolute variations
into one of bounding relative variations, or mismatch.  One approach to reduce
mismatch is to rely on fixed ratios between devices (e.g., in a current
mirror), as the mismatch in ratios is more controllable when nearby devices
experience similar variations (e.g., due to systematic or spatially correlated
effects).  A second approach involves the use of differential structures (e.g.,
differential pairs) for mismatch reduction.  In a typical CMOS process, the
absolute value of device parameters for transistors, capacitors, and resistors
can vary by 20\%, while the requirements on ratio mismatch may be within
0.1\%~\cite{Naiknaware99}.  

Traditionally, these constraints are extracted manually by circuit designers,
relying on expert knowledge.  Automated constraint extraction is one of the
chief bottlenecks~\cite{Scheible2015} to full automation, despite recent
progress in analog automation~\cite{align}.  Manual constraint extraction
methods, which rely on ``designer intent'' and years of experience, are
difficult to translate to an algorithmic methodology.

One class of prior methods is based on sensitivity analysis: Malavasi {\em et
al.}~\cite{Malavasi96} identify matching requirements between nodes with
similar sensitivities.  However, sensitivity analysis is computationally
intensive, especially for nonlinear circuits that require large-change
sensitivities.
%This method requires testbenches and excitations to be specified: these are
%typically provided at the design level but may not be available for internal
%blocks within a larger hierarchical design.  
A second class of methods is topology-based: Eick {\em et al.}~\cite{Eick10}
uses a building block based approach using a signal flow graph method to
extract the symmetries.  While this method is computationally tractable and is
effective on simpler symmetries, it is noted in~\cite{Eick10} that it does not
handle more complex symmetries such as those that are hierarchically nested.  A
third class of recent methods is spectrally based: Kunal {\em et
al.}~\cite{gana} employ graph convolutional networks to identify structures
within graphs and then identify symmetries only using traversal methods similar
to~\cite{Eick10}; Liu {\em et al.}~\cite{Liu20} use spectral methods to solve a
related problem of identifying symmetric circuits at the system level.  

\begin{figure}[bht]
\vspace*{-0.1in}
\centering
\includegraphics[width=0.5\textwidth]{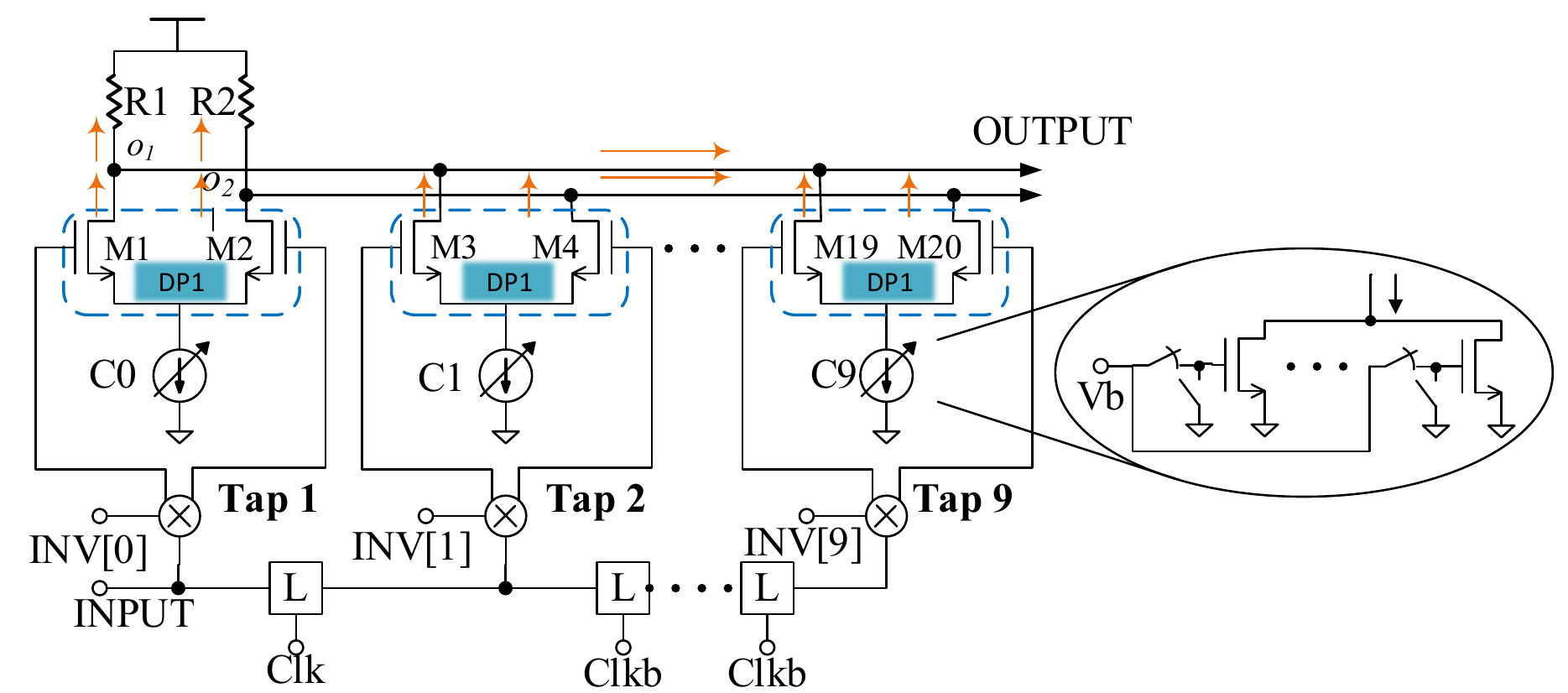}
\vspace*{-0.2in}
\caption{Schematic of an FIR equalizer \cite{Wong06}.}
\label{fig:FIR}
\vspace*{-0.2in}
\end{figure}

No spectral method addresses the full problem of hierarchical constraint
generation in analog circuits with multiple symmetry lines.  Moreover, prior
approaches have only been applied to designs with a small number of blocks,
with relatively simpler symmetries.  To illustrate the complexity that
must be addressed to solve the full problem, we consider the schematic of the
FIR equalizer shown in Fig.~\ref{fig:FIR}, in which any mismatch between
differential pair transistors \{M1-M2\},\{M3-M4\}, $\cdots$, \{M19-M20\}
associated with the taps can result in a gain error. Moreover, the $V_{ds}$
across current sources C0, $\cdots$, C9 should be the same to maintain similar
overdrive current.  A set of symmetry constraints, along multiple axes of
symmetry, must be detected between the transistors in the differential pairs;
the current sources must be self symmetric; each tap and its outputs must be
symmetrically laid out with respect to resistors R1 and R2.  The differential
pairs also have symmetry requirements, and the current driver (shown in the
inset for C9) requires ratioed structures with common-centroid layout for
mismatch reduction.

A complicating factor is that the variable current sources, C0, $\cdots$, C9 may
not be identical: in~\cite{Wong06}, the first 4 taps use 7-bit current-steering
DACs, while the rest employ 5-bit DACs.  The bias voltage is the same for all
taps: thus, despite the small difference in topology, the placement and routing
must be matched. To the best of our knowledge, no existing technique addresses
this problem of detecting symmetries between approximately identical analog
blocks.

The requirements for a methodology that identifies symmetries in a netlist are:
(1)~Speed and scalability to large circuits;
(2)~Ability to identify constraints hierarchically;
(3)~Generality and applicability to a wide range of circuits;
(4)~Capability of identifying multiple axes of symmetry;
(5)~Capacity to identify symmetries between blocks that are approximately
similar and need matching.
Our work all requirements and has the following key features:
\begin{itemize}
\item
It hierarchically handles multiple symmetry levels using a graph-based
framework, and identifies array structures.
\item
It invokes a matching algorithm based on exact or approximate matching:
the latter is based on finding graph edit distances, and employs a graph neural
network (GNN) to detect matching in structures such as Fig.~\ref{fig:FIR}.
\item
It demonstrates solutions on a range of design types, ranging from low-frequency
analog to wireless designs.
\end{itemize} 
Open-source software for this algorithm is available at~\cite{ALIGN-symmetry}.

\ignore{
Moreover, spectral matching works only
on the structure of the graph, and does not inherently embed node features into
the comparison, as we will do in our machine learning based approach.
}

%% file: sec/2-method.tex
\section{Our hierarchical approach}
\label{sec:symmetry}

\subsection{Graph representation and preprocessing}

\noindent
Inspired by~\cite{SubGemini}, we represent a circuit netlist as an undirected
bipartite graph $G(V,E)$.  The set of vertices $V$ can be partitioned into two
subsets, $V_e$, corresponding to the elements
(transistors/passives/ hierarchical blocks) in the netlist, and $V_n$, the set
of nets.  For each element $e$ corresponding to vertex $v_e \in V_e$, 
if net $n$ is incident on $e$, then the net vertex $v_n \in V_n$ is connected
to $v_e$ by an edge in $E$.  There are no edges between two elements in $V_e$,
or two nets in $V_n$, and therefore the graph is bipartite.  Edges to
multiterminal elements are {\em labeled} to indicate which terminal connects to
a net vertex, e.g., each edge from a transistor node is assigned a three-bit
label ($l_gl_sl_d$) marking a gate, source, or drain connection.  We traverse
this graph to hierarchically identify structural symmetries.

\begin{figure}[hbt]
\vspace*{-0.1in}
\centering
\includegraphics[width=0.42\textwidth]{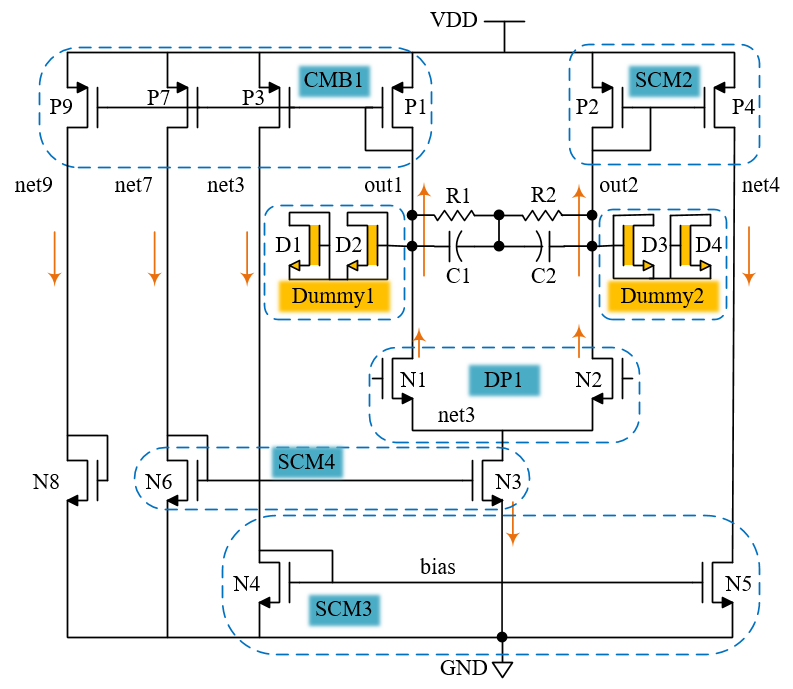}
\vspace*{-0.1in}
\caption{An OTA circuit used to illustrate our approach.}
\label{fig:symm_dp}
\vspace*{-0.25in}
\end{figure}

Initially, the graph is preprocessed using a traversal to identify the lowest
level of building blocks, similar to library building blocks in~\cite{Eick10}
or primitives in~\cite{gana}.  Primitives are composed from a few elements that
form the basic building blocks of an analog circuit, and are specified by the
user in a library.  These are typically simple structures, e.g., passives or
collections of a small number of transistors such as differential pairs,
current mirrors, cascoded structures, or level shifters.  For example, in the
circuit in Fig.~\ref{fig:symm_dp}, four current mirrors -- a current mirror
bank (CMB1) and three single current mirrors (SCM2--SCM4), and a differential
pair (DP) are mapped to library primitives.  
%For example, in a differential pair (DP), both transistors should be placed
%symmetrically, the external tap to the connected source terminal should be at
%the center, and the drain connections should be routed symmetrically.
The elements in a primitive are collapsed into supernode vertices with labeled
ports (net vertices) so that like ports of like vertices can be recognized for
symmetry.  For example, in a DP, the two transistor drain nodes are marked as
symmetric and the source node is labeled so that it can be used in higher-level
symmetry detection.  Such primitive-level symmetry constraints are passed to
the next level of hierarchy.

\subsection{Symmetry detection algorithm}
\label{sec:symmdet}

\noindent
We propose a bottom-up approach that hierarchically detects symmetry
constraints. At each level of hierarchy, symmetrical net information is passed
to higher levels using ports whose information is used to identify matching at
those levels.  

After primitive blocks are identified using graph-based
techniques~\cite{Eick10,gana} and primitive-level symmetries are identified,
a search begins from all {\em potentially symmetric} pairs of node vertices
$(s_1,s_2)$, e.g., transistor drain nodes in a DP, or corresponding nodes of an
SCM and CMB.  For primitives CMB1 and SCM2 in Fig.~\ref{fig:symm_dp}, out1 could
match with out2, and net4 could match with \{net3, net7, net9\}.  Therefore, the
candidate choices for $(s_1, s_2)$ are: (out1, out2), (net3,
net4), (net7, net4) and (net9, net4).  Similarly, DP1 leads to the candidate
(out1,out2), also identified by CMB1/SCM2.

\input sec/2-symmetry_algorithm.tex

The overall algorithm for graph-based symmetry detection operates recursively
and is described in Algorithm~\ref{alg:symm}.  It is initially invoked with set
$(s_1,s_2)$ corresponding to all potential symmetry points from the primitive
set.  All vertices in the graph are marked as unexplored, except for supply
and ground; the algorithm continues traversing the circuit graph and terminates
when no unexplored node can be reached.  The algorithm proceeds as follows: 

\noindent
{\bf Neighbor list}
For each pair of candidate points $(s_1,s_2)$ provided as an input to the
algorithm, we invoke Neighbors(${\cal N}$), which returns the unvisited
neighbors of node ${\cal N}$ (line~\ref{a1:nbrs}).  For transistors, these
neighbors correspond to source/drain-connected vertices. For example, in
Fig.~\ref{fig:symm_dp}, starting from the ports of DP1 (and so DP1 is marked as
visited), for $(s_1, s_2) =$ (out1, out2), the neighbor sets are ${\cal N}_1 =$
\{R1, C1, D1, D2, CMB1/out1\} and ${\cal N}_2 = $ \{R2, C2, D3, D4,
SCM2/out2\}.  (Note that for CMB1 and SCM2, we annotate the node vertices with
the matching port.)

\noindent
{\bf End case detection}
A recursive search is then carried out from the above list of neighbors, with
the end case occurring when $s_1$ and $s_2$ have no unexplored neighbors,
(line~\ref{a1:endrecursion}).

\noindent
{\bf Finding groups of pairs} 
Next, MatchPair detects matches between the vertices of ${\cal N}_1$ and ${\cal
N}_2$, returning ``group\_of\_pairs,'' a set of all matching vertex pairs.  For
example, in Fig.~\ref{fig:symm_dp}, starting from DP1, for $(s_1, s_2) =$
(out1, out2), \\
%\begin{center}
%\text{group\_of\_pairs} =
%\{(D1, D3), (D1, D4), (D2, D3), (D2, D4), \\
%\hspace{0.2in}(R1, R2), (C1, C2), (CMB1/out1, SCM2/out2)\}
%\end{center}
\makebox[3.3in]{group\_of\_pairs = \{(D1, D3), (D1, D4), (D2, D3), (D2, D4),} \\
\makebox[3.3in]{(R1, R2), (C1, C2), (CMB1/out1, SCM2/out2)\}} \\
Thus, from the cross-product of ${\cal N}_1$ and ${\cal N}_2$, this list
eliminates pairs that do not match, e.g., (R1,C1).  When FindSymmetricPairs
is recursively called, it traverses unvisited neighbors of these elements,
e.g., when $(s_1,s_2)=$ (CMB1/out1, SCM2/out2), there are further recursive
calls to match (net3, net4), (net7, net4), and (net9, net4).

The MatchPair function can detect two types of matches:\\
$\bullet$ An {\em exact match}, e.g., a two-terminal element such as a resistor,
represented by ${\cal N}_A$, matches a resistor of identical value, represented
by ${\cal N}_B$.  For a multiterminal element such as transistors, edge labels
are also considered in pronouncing a match.
\\
$\bullet$ An {\em approximate match}, i.e., nonidentical structures to be
matched.\\
In Section~\ref{sec:SimGNN}, we show how a neural network is used to find
the graph edit distance (GED) to predict matches. If the GED is zero, the match
is exact, and if it is small, the match is approximate.
%The algorithm for detecting approximate matches is described in Section~\ref{sec:SimGNN}.  

\noindent
{\bf Processing matches}
Lines~\ref{a1:onepair_begin} -- \ref{a1:onepair_end} consider the case where a
single pair $(p_1,p_2)$ is detected. This means there is a single axis of
symmetry for the pair.  If $p_1$ and $p_2$ are identical, there is a
self-symmetry that begins a new axis of symmetry. Otherwise we have a symmetric
node pair that continues the previous axis; any matching constraints
propagated from inside the block are added to $P$.

For example, starting from CMB1 and SCM2, searching from corresponding ports
$(s_1,s_2) =$ (net3, net4), a unique pair (SCM3/out1, SCM3/out2) is found.
Since these ports are symmetric, $p_1 = p_2$, and the new axis of symmetry lies
at the center of SCM3.  In the next recursive call, since the only unvisited
neighbor of SCM3 is ground, the symmetric axis involving DP1, CMB1, SCM2, SCM3
is complete.

Lines~\ref{a1:multipair_begin} -- \ref{a1:multipair_end} consider cases
where more than one pair is matched: this may lead to $>1$ axis of symmetry if
multiple elements of the same type are matched, e.g., in Fig.~\ref{fig:FIR}.
In lines~\ref{a1:validgroups_begin} -- \ref{a1:validgroups_end}, the
exploration continues recursively from the matched pair until no further
neighbors match.  The valid matching paths, eliminating the nonconverging
paths, are stored in valid\_groups.

If valid\_groups is a singleton, it is added to $P$; else, the multiple matches
correspond to an array of matching elements, rooted at $s_1$ and $s_2$. Each
array is recognized as a hierarchical block, and the matched arrays are added
to $P$.  Array generation is performed by CreateArray($G, s$)
(lines~\ref{a1:Array1}--\ref{a1:Array2}) by collecting a set of repeated
structures connected to a node $s$ of a graph $G$.  For the matches for $(s_1,
s_2) =$ (out1, out2) listed above, D1 and D2 each match with D3 and D4;
therefore, D1+D2 are grouped into array Dummy1, and D3+D4 into Dummy2. Matching
constraints are created between\\
\makebox[3.5in]{(Dummy1, Dummy2), (R1, R2), (C1, C2)}\\
\makebox[3.5in]{(CMB1/out1, SCM2/out2)}
%Array generation is performed by {\bf CreateArray($G, s$)}, which collects a
%set of repeated structures connected to a node $s$ of a graph $G$ of the
%network to create an array.

A key contribution of this algorithm is its ability to build symmetry hierarchies,
as shown in the case with (Dummy1, Dummy2) above.  It could be argued that
this simple illustrative case could be solved by defining a group of two dummy
transistors as a primitive; however, the algorithm is general enough to handle
more complex scenarios that other existing approaches cannot process.
As an example, consider the FIR equalizer in Fig.~\ref{fig:FIR} with multiple
symmetries and array structures, as described in Section~\ref{sec:intro}.  If
$(s_1, s_2) = (o_1, o_2)$, the two nodes from the differential output,
MatchPair would first detect the multiple matches corresponding to the DP.
This would then be extended to a larger structure in valid\_groups by also
including the current source and XOR, where the current source is considered a
match based on the approximate matching scheme to be described in
Section~\ref{sec:SimGNN}.  This combined structure, (DP + current source + XOR),
is assembled into an array.

Finally, Line~\ref{a1:nopair} discards $P$ if no match is found, and the search
from $(s_1, s_2)$ terminates.

%% file: sec/2-symmetry_algorithm.tex
\begin{algorithm}
{\footnotesize
\begin{algorithmic}[1]

    \State {\bf Function} 
    \Function{FindSymmetricPairs }{($G$, ($s_1$, $s_2$), $P$) }	
    \State {{\bf Input}: simplified circuit graph ($G$), start points ($s_1$,$s_2$)}
    \State {{\bf Output}: list of match pairs ($P$)}
	    \State ${\cal N}_1 =$ Neighbors($s_1$) ;
	    ${\cal N}_2 =$ Neighbors($s_2$) \label{a1:nbrs}
	    \IF {${\cal N}_1 == {\cal N}_2 ==$ NULL} \label{a1:endrecursion}
	    	\textit{// End of recursion}
	    	\State {\bf Return}
	    \ENDIF
		\State group\_of\_pairs $=$ MatchPair(${\cal N}_1,{\cal N}_2$)
		\IF {length(group\_of\_pairs) $==1$} \label{a1:onepair_begin}
			%\State	\textit{// Single axis of symmetry}
			\textit{// Single axis of symmetry}
			\State pairs $=$ group\_of\_pairs[0]
			\FOR {$(p_1,p_2) \in$ pairs}
			\State $P.add (p_1,p_2)$
	    		\IF { $p_1 == p_2$ }
					%\State	\textit{// Self-symmetric nodes; New axis of symmetry}
					\textit{// Self-symmetric nodes; New axis of symmetry}
					\State FindSymmetricPairs($G ,(p_1,p_1), P.copy()$)
				\ELSE
					%\State \textit{// Symmetric node pairs}
					\textit{// Symmetric node pairs}
					\State FindSymmetricPairs($G, (p_1,p_2),P$)	
				\ENDIF
			\ENDFOR \label{a1:onepair_end}
		\ELSIF {length(group\_of\_pairs) $>1$} \label{a1:multipair_begin}
			%\State\textit{// Multiple possible axes of symmetry}
			\textit{// Multiple possible axes of symmetry}
			\State valid\_groups $=$ NULL \label{a1:validgroups_begin}
			\FOR {pairs $\in$ group\_of\_pairs}
				%\State  $P_{new}$ = NULL
				\FOR {$(p_1,p_2) \in$ pairs}
					\State $P_{new} = (p_1,p_2)$
					\State FindSymmetricPairs $(G, (p_1,p_2), P_{new})$
					\IF {$P_{new}$} valid\_groups.add$(P_{new})$
					\ENDIF
				\ENDFOR
			\ENDFOR \label{a1:validgroups_end}
			\IF {length(valid\_groups) $== 1$}
				\State $P.add$(valid\_groups[0])
				\State FindSymmetricPairs($G$, valid\_groups[0], $P$)
			\ELSIF {length(valid\_groups) $> 1$}
				%\State \textit{// Multiple parallel symmetrical paths}
				\textit{// Multiple parallel symmetrical paths}
				\State $A_1=$ CreateArray($G, s_1$) \label{a1:Array1}
				\State $A_2=$ CreateArray($G, s_2$) \label{a1:Array2}
				\State $P.add (A_1,A_2)$
				\State FindSymmetricPairs($G, (A_1, A_2), P$)
			\ELSE
				%\State \textit{Nonsymmetrical start-points}
				\State {$P$ = NULL}
			\ENDIF \label{a1:multipair_end}
		\ELSE
			%\State \textit{Nonsymmetrical start-points}
			\State {$P$ = NULL} \label{a1:nopair}
		\ENDIF
	\EndFunction

\end{algorithmic}
}
\caption{Hierarchical symmetry detection algorithm}
\label{alg:symm}
\end{algorithm}

%% file: sec/3-SimGNN.tex
\section{Error tolerant matching}
\label{sec:SimGNN}

\subsection{Problem formulation}

\begin{figure}[hbt]
\centering
\includegraphics[width=3.3in]{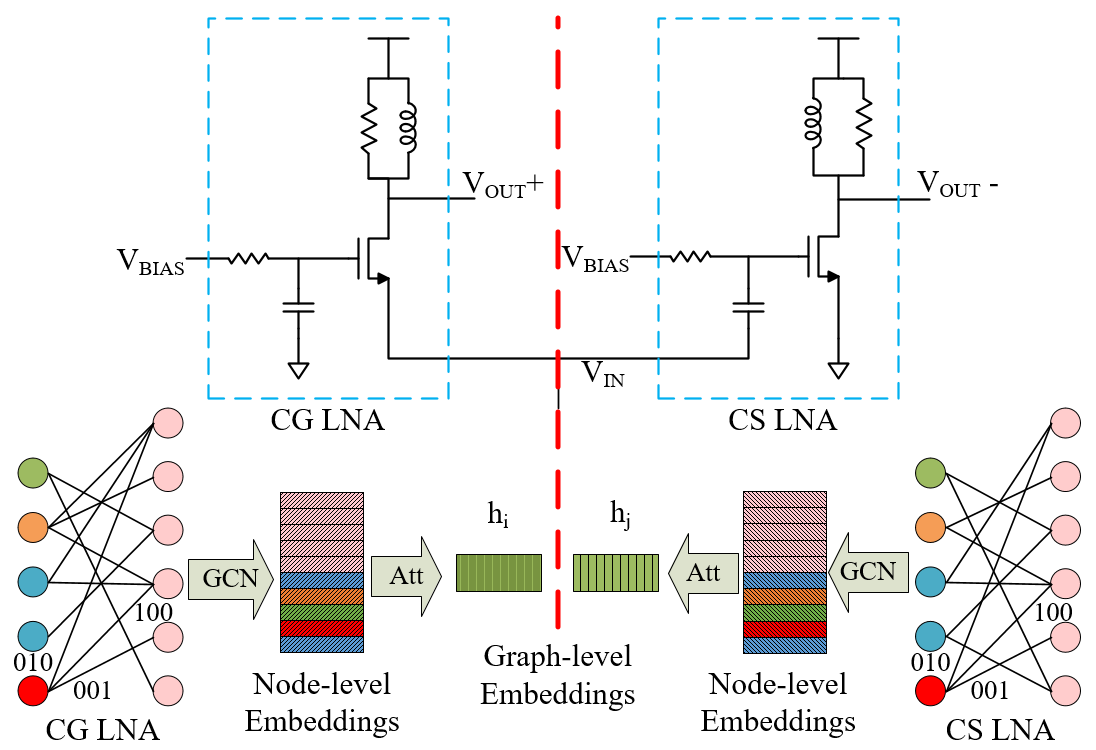}
\caption{Example showing graph embedding for common gate low noise amplifier (CG LNA) and common source LNA (CS LNA) in noise cancellation LNA.}
\vspace*{-0.1in}
\label{fig:ged_example}
%\vspace*{-0.1in}
\end{figure}

\noindent
Fig.~\ref{fig:FIR} had shown an example of the need for approximate matching,
with a different numbers of bits being used in different taps of an equalizer:
despite difference in the topology of the tap control bits, the circuit
requires matching between taps for optimal performance.
Fig.~\ref{fig:ged_example} illustrates another example that shows the matching
requirement between a common-gate low-noise amplifier (CG-LNA) and a
common-source LNA (CS-LNA) in a noise cancellation LNA.  The two sides have a
small difference in topology: the transistor source is connected to the
capacitor terminal at left, but to ground at right.  

These examples show that matching requirements are more complex than the
simpler test cases typically handled in academic papers, and that matching is
frequently required between parts that are similar but not identical.  In fact,
production analog designs use multiple techniques such as asymmetric dummy
transistors in performance-critical parts~\cite{Lin12}, noise cancellation
circuits~\cite{Abidi05,Bruccoleri04}, trim bits to handle noise and
testing~\cite{Moreland,Carley89}, and different device sizes for handling
multiple bands in phased array systems~\cite{ehyaie2011}.  

This implies that the MatchPair function that detects symmetries must allow for
minor changes in circuit topology.  This is the inexact graph matching problem,
and we map this to the Graph Edit Distance (GED) problem.  The GED a measure of
similarity between two graphs $G_1$ and $G_2$: given a set of graph edit
operations (insertion, deletion, vertex/edge relabeling), the GED is a metric
of the number of edit operations required to translate $G_1$ to $G_2$.  

Let graphs $G_1$ and $G_2$ represent, respectively, the CS-LNA and the CG-LNA,
as shown in Fig.~\ref{fig:ged_example}, with element vertices at left and net
vertices at right.  To transform $G_1$ to $G_2$, four edits are needed (i.e.,
GED = 4):
(1)~Deletion of two edges in $G_1$: (capacitor element, ground net) and
(transistor element (source label), $V_{IN}$ net), and (2)~Addition of two
edges in $G_1$: (capacitor element, $V_{IN}$ net)
and (transistor element (source label), ground net).

We calculate the similarity between two subblocks by comparing graph embedding
of the two graphs. If the similarity is within a bound, the MatchPair function
in Algorithm~\ref{alg:symm} returns a match.

\subsection{Graph neural network formulation}

The GED problem is NP-hard~\cite{Zeng09}, implying that an exact solution is
computationally expensive.  This work uses a neural network that transforms
the original NP-hard problem to a learning problem~\cite{bai18} for computing
graph similarity.  

\begin{figure}[t]
\centering
\includegraphics[width=3.3in]{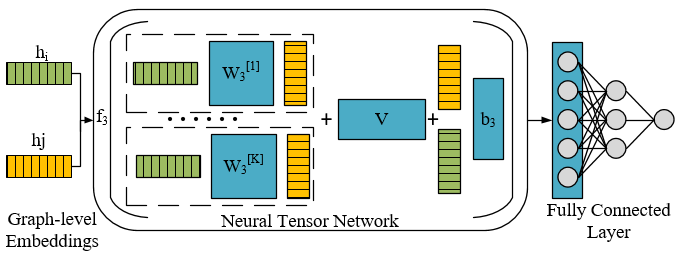}
\vspace*{-0.1in}
\caption{GED prediction based on graph embeddings~\cite{bai18}.}
\label{fig:network}
\vspace*{-0.2in}
\end{figure}

The method works in four steps: first, each node in the graph is converted to a
node-level embedding vector; next, these embedding vectors are used to create a
{\bf graph-level embedding of dimension $d$}.  The lower half of
Fig.~\ref{fig:ged_example} illustrates these two steps for the graphs for the
CG-LNA and CS-LNA.  For each subblock in the circuit, these steps need to be
carried out once, and the graph embeddings are stored for matching any two
pairs of subblocks in later stages.  The computational complexity of these two
steps is linear in the number of nodes in the graph.  The last two steps are
shown in Fig.~\ref{fig:network}. In the third step, the graph-level embeddings
from the second step for two candidate graphs are fed to a trained neural
tensor network that generates a similarity matrix between the graphs. The
fourth step then processes this matrix using a fully connected neural network
to yield a single score.  This matching method for two subblocks uses the
previously stored graph embeddings instead of the full subblock graphs.  The
complexity of these two steps is quadratic in $d$, where $d$ is bounded by a
small constant in practice, the procedure is computationally inexpensive as
compared to an exact GED computational complexity which is exponential in the
number of nodes of the graphs involved.

\noindent
\underline{\bf Node embedding stage:}
This stage transforms each node of a graph into a vector, encoding $d$ node
features and neighborhood information in a manner that is
representation-invariant.  We use neighbor feature aggregation based on
a three-layer graph convolutional network (GCN)~\cite{Kipf2017} to obtain the
node embedding.  The output $X_{l+1} \in \mathbb{R}^{N \times d}$ in layer $l+1$ from
the value in layer $l$ as: 
\begin{equation}
X^{l+1} = \text{ReLU} ( \hat{D}^{-1/2} \hat{A} \hat{D}^{-1/2} X^l W_1^l)
\end{equation}
where ReLU$(x)=\max(0,x)$  is the activation function, $\hat{A} = A + I_N \in
\mathbb{R}^{N \times N}$ is the adjacency matrix of an undirected graph with added
connections for each vertex to itself, $\hat{D} \in \mathbb{R}^{N \times N}$ is the
diagonal matrix of $\hat{A}$, and $W_1^l \in \mathbb{R}^{d^l \times d^{l+1}}$ are
trainable weights for layer $l$. 

The GCN output, $X^{3}$, is the node embedding matrix, $X$.  The
$n^{\rm th}$ row of $X$, ${\bf X}_n^T \in \mathbb{R}^d$, is the embedding of node $n$.

\noindent
\underline{\bf Graph embedding stage:}
For each graph to be compared, we now produce an embedding using the
attention-based aggregation of node embeddings generated in the previous stage.

We first compute a global context, ${\bf c}^T \in \mathbb{R}^d$, for the graph,
computed as weighted sum of node mbedding vector averages,
%These graph embeddings are stored as node features, representing information of
%the subblock graph. 
%A weighted sum of the average of node embeddings is taken,
followed by a nonlinear transformation, using trainable weights $W_2 \in
\mathbb{R}^{d \times d}$:
\begin{equation}
{\bf c} = \tanh \left [ \frac{\sum_{m=1}^{N} X_m}{N} W_2 \right ]
\label{eq:graph_embed:c}
\end{equation}
Here, $N$ is the number of nodes in the graph.  Next, we use an attention
mechanism to allow the model weights to focus on important parts of
circuit, guided by the GED similarity metric. We empirically observe that nodes
with high degree, and nodes forming special structures such as loops, get higher
attention weights: this is because high-degree nodes receive contributions from
a larger number of neighbors.  The graph embedding ${\bf h} \in \mathbb{R}^d$
is given by:
\begin{align}
\textstyle
{\bf h} = \sum_{n=1}^N \sigma ( {\bf X}_n {\bf c}^T ) {\bf X}_n 
\label{eq:graph_embed}
\end{align}
where $\sigma(x) = 1/(1+\exp(-x))$ is the sigmoid function.
%, and $c$ is given by \eqref{eq:graph_embed:c}, applied elementwise to ${\bf h}$.

\noindent
\underline{\bf Neural Tensor Network stage:}
Next, the relationship between two graph embeddings, ${\bf h}_i, {\bf h}_j \in
\mathbb{R}^d$, is measured using Neural Tensor Networks
(NTNs)~\cite{reasoning2013} as:
\begin{equation}
g({\bf h}_i,{\bf h}_j) = \text{ReLU} ( {\bf h}_i^T W_3^{[1:K]} {\bf h}_j +
				V [{\bf h}_i {\bf h}_j]^T + b )
\end{equation}
where $K$ is a hyperparameter related to number of slices in the tensor, which
controls the number of similarity scores produced by the model, $W_3^{[1:K]} \in
\mathbb{R}^{d \times d \times K}$ is a weight tensor, $V \in \mathbb{R}^{K \times 2d}$ is a
weight vector, and $b \in \mathbb{R}^K$ is a bias vector.

\noindent
\underline{\bf Graph similarity score computation stage:} 
The final step reduces the similarity scores in previous stage using a
two-layer fully-connected neural network to provide a predicted
similarity score $PS$. To train this network, the final score is compared
against the ground truth GED score $GS$ using the mean square error loss:
\begin{equation}
\textstyle
L = \frac{1}{|S|} \sum_{(i,j) \in S} (PS_{ij} - GS_{ij})
\label{eq:graphsimilarityscore}
\end{equation}
where $S$ is the set of training graph pairs.

\subsection{Training and hyperparameter tuning}

\begin{figure}[hbt]
\centering
\includegraphics[width=0.4\textwidth]{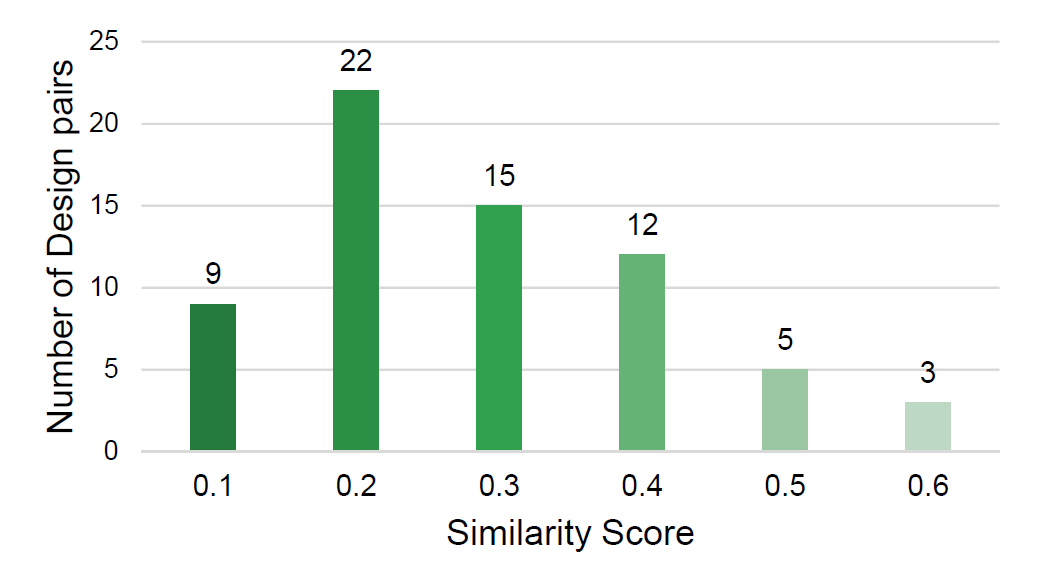}
\vspace*{-0.1in}
\caption{Distribution of the results of training into bins to obtain the similarity
score used in Eq.~\eqref{eq:graphsimilarityscore}.}
\label{fig:data}
\vspace*{-0.1in}
\end{figure}

\noindent
We have trained our network on 79 pairs of analog designs, where each pair has
a small difference in topologies.  Examples in our training set include
single-ended vs. differential OTAs, multiple common-gate vs. common-source
LNAs, OTAs with dummies, and arrays of current mirrors of different
sizes.  For each pair, the ground truth GED was computed using the algorithm in
~\cite{zeina15}, and a similarity metric between graphs $G_1$ and
$G_2$ was defined as:
\begin{align}
\text{dist} (G_1, G_2) =
\frac{\text{GED}(G_1,G_2)}{(|G_{1v}|+|G_{1e}|+|G_{2v}|+|G_{2e}|)}
\end{align}
where $|G_{iv}|$ ($|G_{ie}|$) is the number of vertices (edges) in $G_i$.  
The denominator normalizes the GED to the size of the graph.
Next, recognizing that the GED score is more qualitative than quantitative
(i.e., accuracy of multiple decimal places does not matter), we divide these
distance scores into bins that define the level of similarity, as illustrated
in Fig.~\ref{fig:data}.  This score is used in
Eq.~\eqref{eq:graphsimilarityscore} both for training and inference, to
quantify the match between candidate pairs.

\begin{figure}[hbt]
\centering
\includegraphics[width=0.4\textwidth]{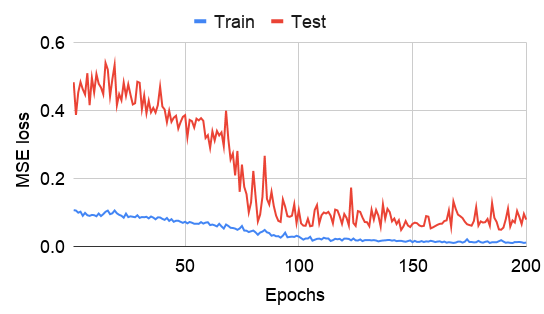}
\vspace*{-0.1in}
\caption{Results on training the network.}
\label{fig:training}
\vspace*{-0.2in}
\end{figure}

\begin{figure}[hbt]
\centering
\includegraphics[width=0.4\textwidth]{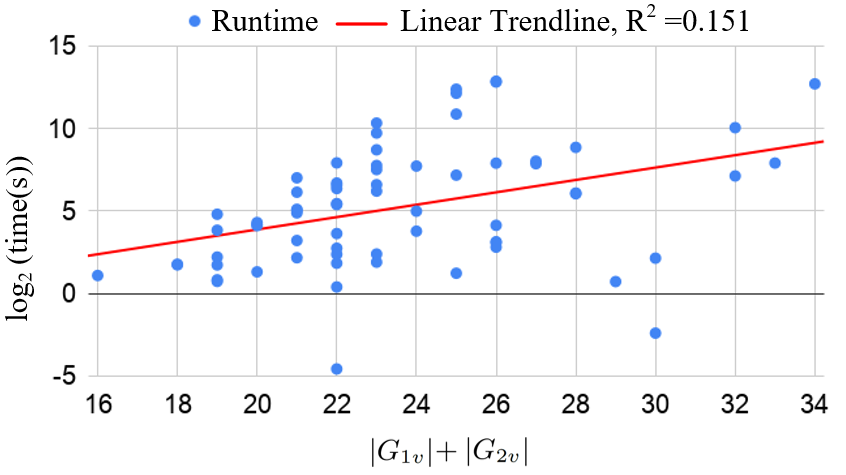}
\vspace*{-0.1in}
\caption{Computational cost of the GED algorithm in~\cite{zeina15}.}
\label{fig:GED_complexity}
\vspace*{-0.1in}
\end{figure}

For a train:test ration of 51:28 among the 79 pairs,
Fig.~\ref{fig:GED_complexity} shows the computation time for calculating
correct GED:  it can be seen to increase exponentially with the graph size.

We have experimented with several options in training with regard to the use of
edge labels: recall that edge labels are used for vertices in the graph that
represent multiterminal elements.  The simplest version ignored the edge labels
and performed matching without labels, while a more complex version required us
to modify the graph embeddings at the GCN stage to enable the use of edge
labels.  We verified during training our model that the use of edge labels is
important for improved accuracy as it provides more information about the
presence of a drain/gate/source connection to a transistor. 

\ignore{
\begin{figure}[hbt]
\centering
\includegraphics[width=0.45\textwidth]{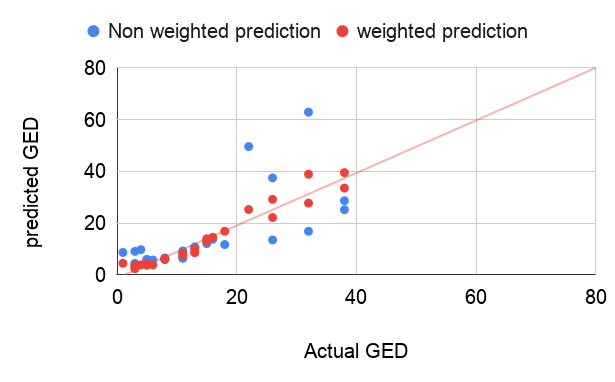}
\caption{Test results on 20 pairs of circuits.}
\label{fig:testing}
\end{figure}
}

\begin{figure}[hbt]
\centering
\includegraphics[width=0.45\textwidth]{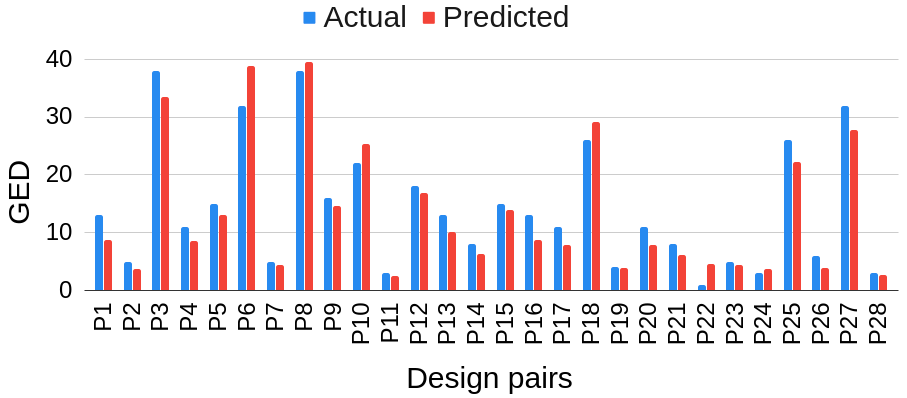}
\vspace*{-0.1in}
\caption{Comparison of GED prediction using our approach and~\cite{zeina15} on
a set of pairs of circuit graphs.}
\label{fig:testing_v2}
\vspace*{-0.1in}
\end{figure}

\ignore{
The difference is illustrated in Fig. \ref{fig:training}, where we see an
improvement in the loss metric when we use edge labels  during training (the
mean square error, $MSE=0.012$) as compared to one without edge labels
($MSE=0.015$). \blueHL{In the scatter plot of predicted GED Fig.~\ref{fig:testing} we can see that the prediction with edge labels is much closer to the $45\deg$ red line showing improved results with edge labels}. 
}

Fig.~\ref{fig:GED_complexity} illustrates the reduction of the loss metric in
successive epochs during training.  For both training and test phases, a steady
reduction in loss is noted.  Fig.~\ref{fig:testing_v2} compares the GED
predicted by our approach with the slower GED calculation from~\cite{zeina15},
and shows a good match.  The relatively small, but noticeable, magnitude of the
difference is reflective of the fact that shows that many elements of the
training set do not exercise a topology difference that involves a
multiterminal element with labels.

Our optimized model is a three-layer GCN with 128 input channels (the number
of channels is halved in each layer), with 8 slices in the NTN, and a fully
connected network with one hidden layer after the NTN.
The trained net uses $d_1 = 64, d_2 = 32, d_3 = d = 16$.

%% file: sec/4-results.tex
\begin{figure}[hbt]
\centering
\begin{subfigure}
\centering
\includegraphics[width=0.40\textwidth]{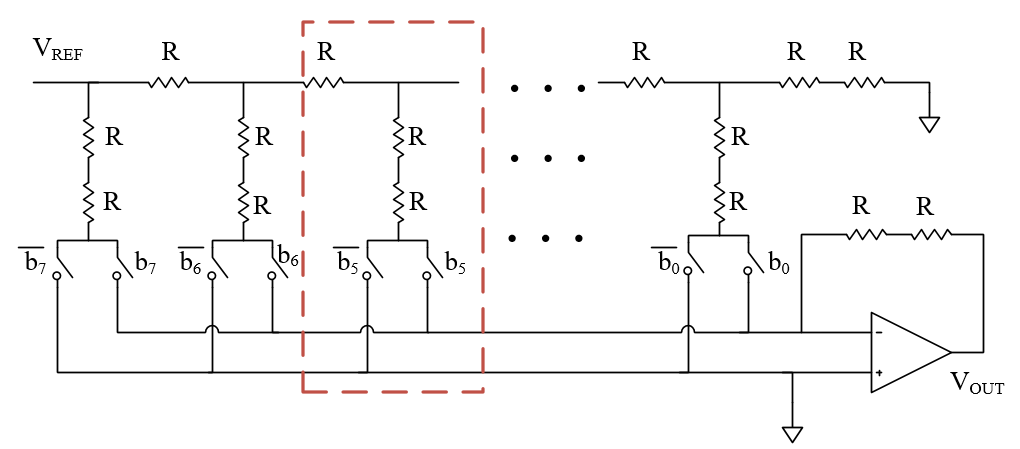}
\label{fig:r2r_sch}
\end{subfigure}
\begin{subfigure}
\centering
\includegraphics[width=0.50\textwidth]{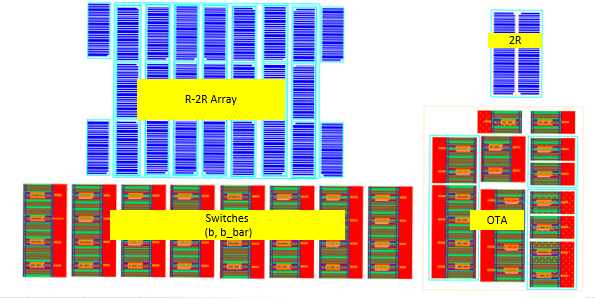}
\label{fig:r2r_lay}
\end{subfigure}
\vspace*{-0.1in}
\caption{(top) Schematic of an R2R DAC with an R2R ladder recognized
using our algorithm.  (bottom) R2R DAC layout.}
\label{fig:r2r}
\vspace*{-0.1in}
\end{figure}

\section{Experimental results}
\label{sec:results}

\noindent
We first present results on three designs that exercise hierarchical
symmetries: the OTA of Fig.~\ref{fig:symm_dp}, an R-2R DAC shown in
Fig.~\ref{fig:r2r}(top), and the FIR equalizer of Fig.~\ref{fig:FIR}. The
characteristics of the graphs of these circuits are summarized in
Table~\ref{tbl:data}.  These circuits contain sufficient complexity to exercise
our method and demonstrate its validity and effectiveness.  Our symmetry
detection algorithm is integrated into the public-domain open-source analog
layout tool, ALIGN~\cite{align}, to generate layouts for these circuits.  For
clarity and to demonstrate the ability of our approach to identify symmetries,
only the placement is shown, without routing.

\input tbl/ckts.tex

\noindent
{\bf R-2R DAC (Fig.~\ref{fig:r2r}(top))}: The main source of nonlinearity is
mismatch between the array resistors.  The area of the resistors in this
circuit is significant as compared to the CMOS devices, and the resistors must
be placed close to each other in a symmetric common-centroid configuration for
matching.  Our algorithm detects an array of the repeating R-2R module (shown
using the dashed rectangle in the figure) and uses multiple instantiations of
this to create an R-2R array.  The OTA is also hierarchically recognized and
extracted into a module.  The switches $b_7-b_0$ and $\bar{b}_7-\bar{b}_0$ are
part of a digital block, and may be placed outside the resistor array.  The
result of the ALIGN-generated layout using our constraint generation
methodology is shown in Fig.~\ref{fig:r2r}(bottom).

\begin{figure}[hbt]
\centering
\includegraphics[width=0.4\textwidth]{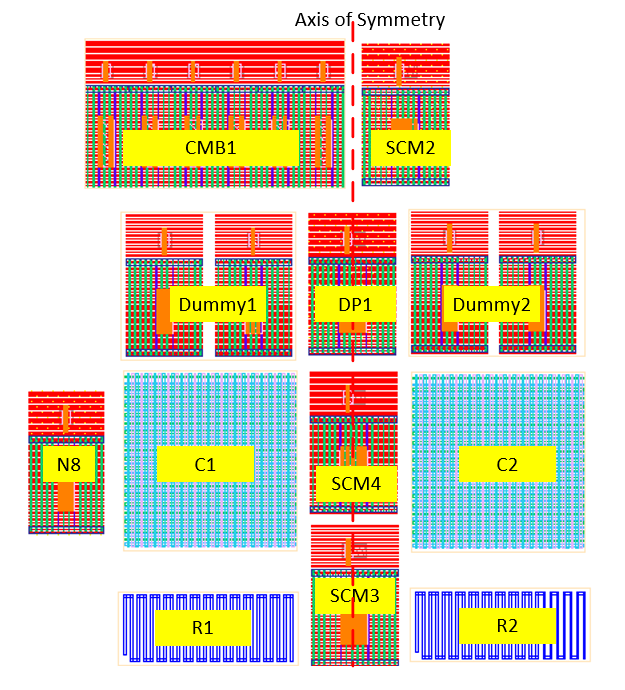}
\vspace*{-0.1in}
\caption{Layout of the OTA in Fig.~\ref{fig:symm_dp}.}
\label{fig:ota_v1}
\vspace*{-0.1in}
\end{figure}

\begin{figure*}[hbt]
\centering
\hspace*{-0.025\textwidth}
\includegraphics[width=0.95\textwidth]{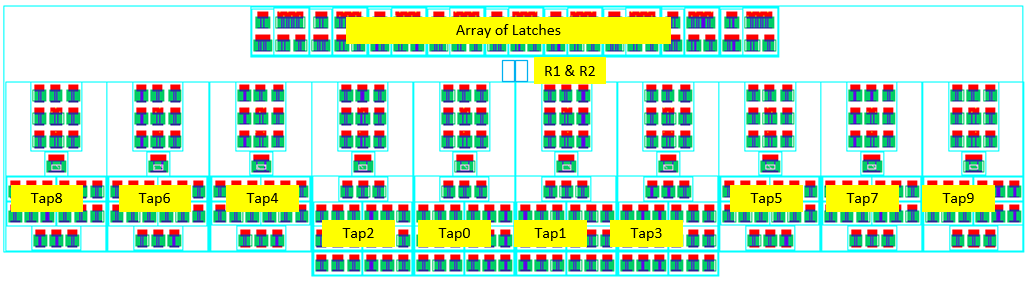}
%\vspace*{-0.2in}
\caption{Layout of the FIR equalizer of Fig.~\ref{fig:FIR}.}
\label{fig:fir_v1}
\vspace*{-0.1in}
\end{figure*}

\begin{figure*}[hb!t]
\centering
%\hspace*{-0.05\textwidth}
%\includegraphics[width=0.55\textwidth]{figures/circuits_final_v3.pdf}
%\includegraphics[width=0.80\textwidth]{figures/circuits_final_v4.pdf}
\includegraphics[width=0.95\textwidth]{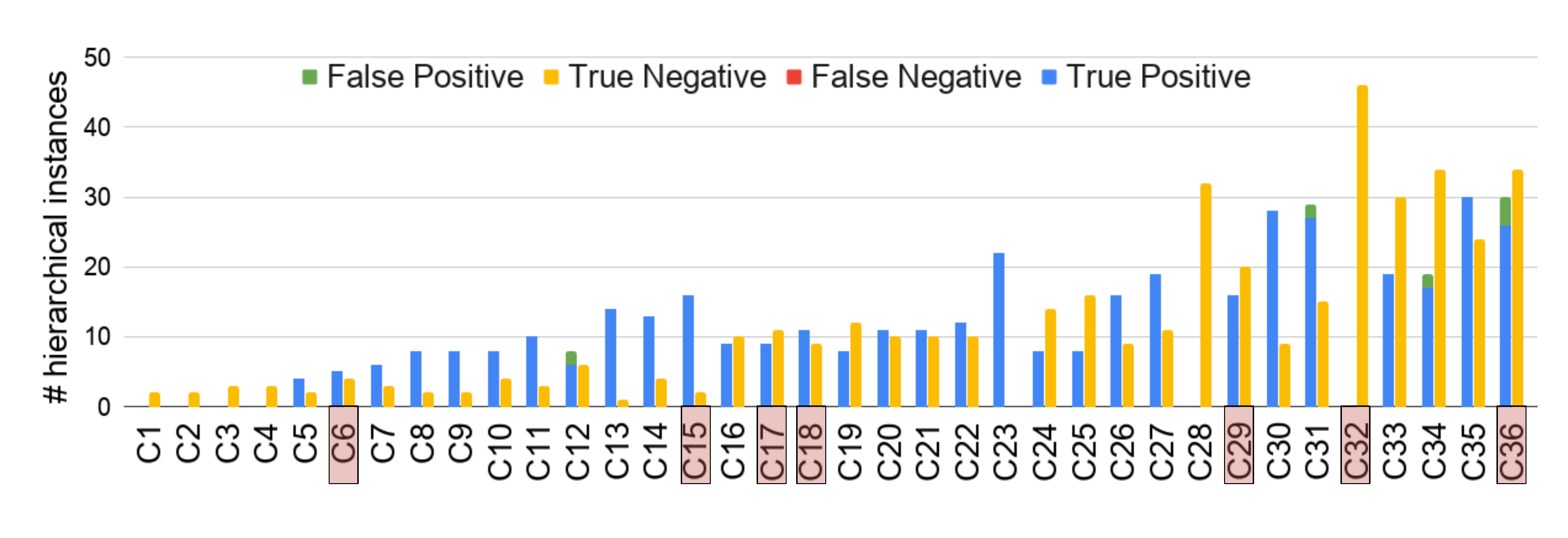}
\vspace*{-0.2in}
\caption{Prediction accuracy of our algorithm, compared with manual verification.}
\label{fig:all_circuits}
\vspace*{-0.1in}
\end{figure*}

\noindent
{\bf OTA circuit (Fig.~\ref{fig:symm_dp})}: Our algorithm detects the
symmetrical axis of the circuit, and the layout about the symmetric line is
shown in Fig.~\ref{fig:ota_v1}.  The dummy blocks, Dummy1 and Dummy2, that were
detected (as described in Section~\ref{sec:symmdet}) must be placed
symmetrically with respect to the differential pair (DP1).  Similarly,
resistors R1 and R2 share the same symmetry axis with DP1, as do capacitors C1
and C2, the current mirrors SCM3 and SCM4.  Transistors P1 and P2 are
symmetric, and the transistors in CMB1 are in common-centroid.

\noindent
{\bf FIR Equalizer (Fig.~\ref{fig:FIR})}: This circuit has ten taps for
equalization, each containing an differential pair, a current mirror DAC, and
CML XOR gate. All blocks in each tap share a common symmetry axis for matching.
The first four taps use a 7-bit current mirror DAC, and the remaining taps have
5-bit current mirror DAC. To achieve better matching, the first four taps are
placed in the center and the remaining taps are placed around these four,
sharing a common symmetry axis. The layout of equalizer, shown in
Fig.~\ref{fig:fir_v1}, meets all these requirements.  This design demonstrates
the detection and use of multiple lines of symmetry in a hierarchical way
within primitives, within each tap, and globally at the block level.

The algorithm presented in this paper has been tested on a wide range of
over a variety of circuits, including OTAs, buffers comparators, VCOs,
analog-to-digital and digital-to-analog converters, and filters.  The circuits
are preprocessed to remove dummy transistors for which all terminals are
connected to supply/ground lines.

Beyond the three circuits discussed earlier, we were able to verify the
correctness of the constraints through manual inspection on 36 circuits.  The
results of this verification, showing true/false positives and true/false
negatives, are summarized in Fig.~\ref{fig:all_circuits}.  For each
hierarchical instance (e.g., device, passive, detected array) in the circuit, a
true positive implies that the instance has a symmetry constraint that is
correctly identified; a negative implies no symmetry constraint.  Symmetrical
instances must be connected using symmetrical nets. For the circuit names
highlighted on the x-axis, our algorithm created a new hierarchy for arrays in
these designs by grouping like elements.  For the smallest circuits, it can be
seen that no symmetries are identified.  From C5 onwards, most circuits have
some symmetry constraints (with the exception of C28, a DC-DC converter, and
C32, an inverter-based VCO); a few circuits (C13, a fully differential
telescopic OTA, and C23, a switched-capacitor filter) have symmetry constraints
involving most/all devices.  Most of the constraints detected by our algorithm
are true positives or true negatives.  No false negatives were detected.

Four circuits have false positives, related to (a) level-shifter structures
connected to the output stages of amplifiers, and (b) dummy structures.  The
former is not harmful because it is connected symmetric current mirror units,
and therefore it is logical to place these symmetrically even though matching
is not required.  The latter is a nonintuitive use of a dummy structure: since
dummies are used for corrections subsequent to first silicon, it is recommended
that any symmetries involving them should be annotated by the designer.

\ignore{
  structures that require weak
matching (we conservatively list these as false positives) or no matching. For
example, in one circuit, analog inverter structures are placed at the output
stage of an OTA, and our algorithm identifies matched transistors within these
structures. These structures are created as trim structures that can allow easy
tuning of the design after a silicon spin.  Enforcing this match does not
result in an incorrect layout, but may create additional constraints.  However,
our layout for this circuit remains compact even with these additional
constraints.  \ignore{ The first class of causes of these are related to
current mirror banks (CMBs), in which a reference current is matched to a set
of output currents.  In the most stringent matching requirement (assumed in our
work), all pairs of currents must be matched: this results in the lowest
input-referred offset performance specification.  In cases where the offset is
not important, it may be sufficient to match However, since our matching
algorithm is purely topological and does not consider performance
specifications, we use a fixed strategy.  Specifically, by default we consider
that matching between all the transistors in CMB is important and we use a
common centroid layout primitive that matches all transistors. A designer may
choose not to use this layout methodology by removing this primitive from the
flow: in such a case, we will detect matching based on symmetrical branch
traversal.

A second class that arises in one specific circuit is a scenario where multiple
}
}

%% file: tbl/ckts.tex
% Please add the following required packages to your document preamble:
% \usepackage{booktabs}
% \usepackage{graphicx}
%\pm 0.76 \pm 2.74  \pm 2.21 \pm 2.69

\begin{table}[hbt]
\caption{Statistics of the graphs for three test designs}
\label{tbl:data}
\vspace{-0.1in}
\centering
%\resizebox{1.00\columnwidth}{!}{%
\begin{tabular}{@{}|c|c|c|@{}}
\hline

Method                 & \#nodes & \#Edges \\ \hline
%All disconnected nodes & 67.32 $\pm$ 0.76 \\ \hline
OTA            & 27 & 34 \\ \hline
R2R ladder            & 116 & 144 \\ \hline
FIR equalizer            & 640  & 1099 \\ \hline

\end{tabular}%
%}
\vspace{-0.1in}
\end{table}

%% file: sec/5-conclusion.tex
\section{Conclusion}

This paper has proposed an approach to handle multiple levels of symmetry
hierarchies, including nested hierarchies, in analog circuits.  
The core algorithm is based on graph traversal through the network graph,
and includes both exact graph-based matching and a novel machine learning based
approximate matching technique using a GCN and a neural tensor network based
GNN.  We validate our results on a variety of designs, demonstrating the
detection of multiple lines of symmetry, hierarchical symmetries, and
common-centroid structure detection.  We show how these guidelines are
transferred for implementation to a layout generation tool that provides high
layout quality.

%\begin{itemize}
%	\item Handles high variability
%	\item Verified on large circuits
%	\item Can be used on industrial design
%	\item need to add more classes and circuits
%\end{itemize}